%% file: main.tex
\documentclass[10pt,twocolumn,letterpaper]{article}

\pdfmapfile{+download35.map}
\usepackage{cvpr}              
\usepackage{graphicx}
\usepackage{amsmath}
\usepackage{amssymb}
\usepackage{booktabs}
\usepackage{bigstrut}
\usepackage{array}
\usepackage{colortbl}
\usepackage[T1]{fontenc}
\usepackage{wrapfig,lipsum}
\usepackage{booktabs}
\usepackage{rotating}
\usepackage{bigstrut}
\usepackage{multirow}
\usepackage{placeins}
\usepackage{gensymb}
\usepackage{xcolor}
\usepackage{soul}
\usepackage{makecell}
\makeatletter  
\@namedef{ver@everyshi.sty}{}
\makeatother
\usepackage{tikz}
\usepackage{yfonts}
\usepackage{graphicx}
\usepackage{import}
\usepackage{pifont} 
\usepackage{appendix}
\usepackage[accsupp]{axessibility}  
\usepackage{hhline}
\usepackage{stfloats}
\usepackage{setspace}
\usepackage{bigstrut}
\usepackage{wasysym}
\usepackage{atbegshi,picture}   
\usepackage{mathtools}
\usepackage[font={footnotesize}]{caption}

\definecolor{cvprblue}{rgb}{0.21,0.49,0.74}
\usepackage[pagebackref,breaklinks,colorlinks,allcolors=cvprblue]{hyperref}
\newcommand{\boldparagraphstart}[1]{\vspace{1pt}\noindent \textbf{#1}}
\newcommand{\xmark}{\text{\ding{55}}}  
\definecolor{darkgreen}{RGB}{0,127,0}
\definecolor{darkred}{RGB}{200,0,0}
\def\greencheckmark{\textcolor{darkgreen}{\checkmark}}
\def\redxmark{\textcolor{darkred}{\xmark}}

\def\authorspace{\quad\quad}

\newcommand*{\affmark}[1][*]{\textsuperscript{#1}}
\newcommand{\thickline}[0]{\Xhline{1.5pt}}

\newenvironment{myitem}{\begin{list}{$\bullet$}
{\setlength{\itemsep}{-0pt}
\setlength{\topsep}{0pt}
\setlength{\labelwidth}{5pt}
\setlength{\leftmargin}{10pt}
\setlength{\parsep}{-0pt}
\setlength{\itemsep}{0pt}
\setlength{\partopsep}{0pt}}}%
{\end{list}}

\begin{document}

\title{FoundationStereo: Zero-Shot Stereo Matching}

\author{Bowen Wen\affmark[] \authorspace Matthew Trepte\affmark[] \authorspace Joseph Aribido\affmark[] \authorspace Jan Kautz\affmark[] 
\\
Orazio Gallo\affmark[] \authorspace Stan Birchfield\affmark[]\\ 
\\
{
\affmark[]NVIDIA}
}

\setlength{\belowdisplayskip}{2pt} 
\setlength{\abovedisplayskip}{2pt} 
\setlength{\belowdisplayshortskip}{2pt} 
\setlength{\abovedisplayshortskip}{2pt}

\twocolumn[{
\renewcommand\twocolumn[1][]{#1}%
\maketitle

\begin{center}
    \centering
     \vspace{-25pt}
     \includegraphics[width=0.99\textwidth]{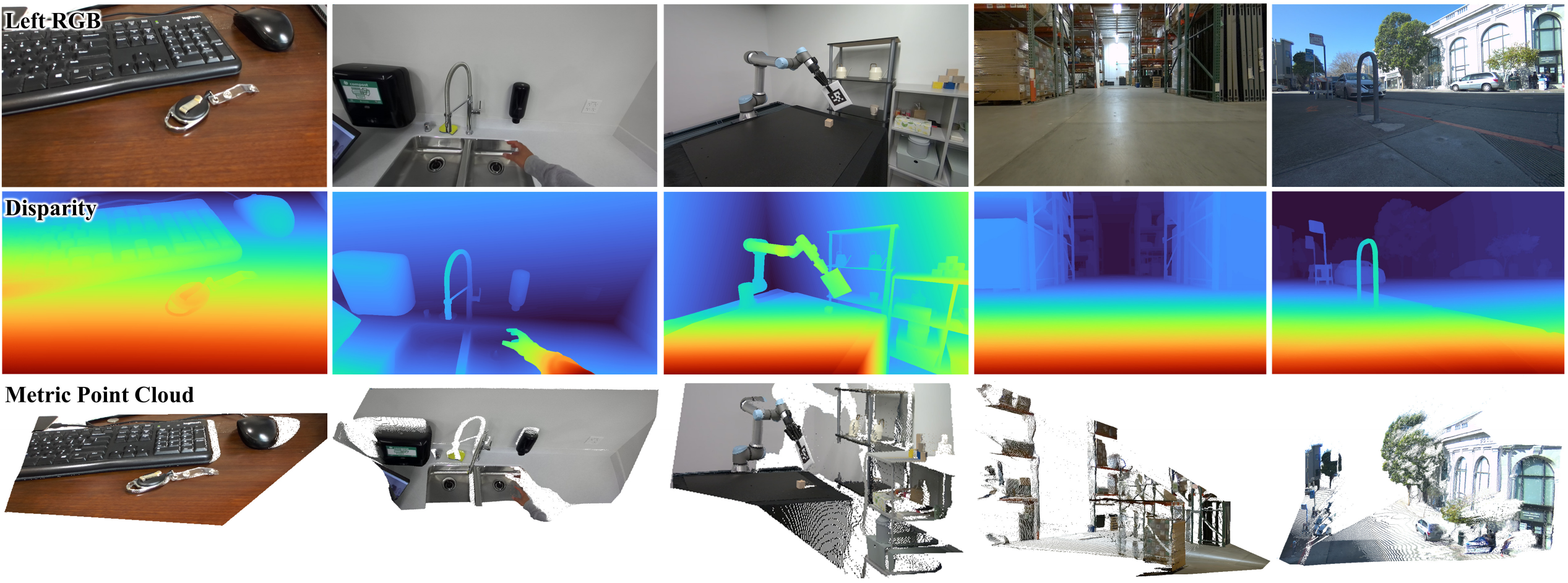}
      \vspace{-0.15in}
    \captionof{figure}{Zero-shot prediction  on in-the-wild images. Our method generalizes to diverse scenarios (indoor / outdoor), objects of challenging properties (textureless / reflective / translucent / thin-structured), complex illuminations (shadow / strong exposure), various viewing perspectives and sensing ranges.
    } \label{fig:intro}
\end{center}%
}]

\begin{abstract}
Tremendous progress has been made in deep stereo matching to excel on benchmark datasets through per-domain fine-tuning. However, achieving strong zero-shot generalization — a hallmark of foundation models in other computer vision tasks — remains challenging for stereo matching. We introduce FoundationStereo, a foundation model for stereo depth estimation designed to achieve strong zero-shot generalization. To this end, we first construct a large-scale (1M stereo pairs) synthetic training dataset featuring large diversity and high photorealism, followed by an automatic self-curation pipeline to remove ambiguous samples. We then design a number of network architecture components to enhance scalability, including a side-tuning feature backbone that adapts rich monocular priors from vision foundation models to mitigate the sim-to-real gap, and long-range context reasoning for effective cost volume filtering. Together, these components lead to strong robustness and accuracy across domains, establishing a new standard in zero-shot stereo depth estimation. Project page: \url{https://nvlabs.github.io/FoundationStereo/}
\end{abstract}
\vspace{-20pt}

\section{Introduction}
\vspace{-5pt}
Since the advent of the first stereo matching algorithm nearly half a century ago~\cite{marrpoggio1976}, we have come a long way. Recent stereo algorithms can achieve amazing results, almost saturating the most challenging benchmarks---thanks to the proliferation of training datasets and advances in deep neural network architectures.
Yet, fine-tuning on the dataset of the target domain is \emph{still} the method of choice to get competitive results.
Given the zero-shot generalization ability shown on other problems within computer vision  via the scaling law~\cite{kirillov2023segment,oquab2023dinov2,yang2024depth,yang2024depthanythingv2}, what prevents stereo matching algorithms from achieving a similar level of generalization?

Leading stereo networks~\cite{xu2020aanet,yang2020waveletstereo,shen2021cfnet,mao2021uasnet,shen2022pcw,chen2023learning}
construct cost volumes from the unary features and leverage 3D CNNs for cost filtering. Refinement-based methods~\cite{lipson2021raft,zhao2023high,tosi2024neural,gong2024learning,crestereo,jing2023uncertainty,selective_stereo,chen2024mocha} iteratively refine the disparity map based on recurrent modules such as Gated Recurrent Units (GRU). Despite their success on public benchmarks under per-domain fine-tuning setup, however, they struggle to gather non-local information to effectively scale to larger datasets.
Other methods~\cite{li2021revisiting,weinzaepfel2023croco} explore transformer architectures for unary feature extraction, while lacking the specialized structure afforded by cost volumes and iterative refinement to achieve high accuracy.

Such limitations have, to date, hindered the development of a stereo network that generalizes well to other domains. While it is true that cross-domain generalization has been explored by some prior works~\cite{zhang2020domain,zhang2025learning,chuah2022itsa,chang2023domain,rao2023masked,liu2022graftnet}, such approaches have not achieved results that are competitive with those obtained by fine-tuning on the target domain, either due to insufficient structure in the network architecture, impoverished training data, or both.
These networks are generally experimented on Scene Flow~\cite{sceneflow2016}, a rather small dataset with only 40K annotated training image pairs.
As a result, none of these methods can be used as an off-the-shelf solution, as opposed to the strong generalizability of vision foundation models that have emerged in other tasks.

To address these limitations, we propose FoundationStereo, a large foundation model for stereo depth estimation that achieves strong zero-shot generalization without per-domain fine-tuning.
We train the network on a large-scale (1M image pairs) high-fidelity synthetic training dataset with high diversity and photorealism. An automatic self-curation pipeline is  developed to eliminate the ambiguous samples that are inevitably introduced during the domain randomized data generation process, improving both the dataset quality and model robustness over iterate updates. 
To mitigate the sim-to-real gap, we propose a side-tuning feature backbone that adapts internet-scale rich priors from DepthAnythingV2~\cite{yang2024depthanythingv2} that is trained on real monocular images to the stereo setup. To effectively leverage these rich monocular priors embedded into the 4D cost volume, we then propose an Attentive Hybrid Cost Volume (AHCF) module,  consisting of 3D Axial-Planar Convolution (APC) filtering that decouples standard 3D convolution into two separate spatial- and disparity-oriented 3D convolutions, enhancing the receptive fields for volume feature aggregation; 
and a Disparity Transformer (DT) that  performs self-attention over the entire disparity space within the cost volume, providing long range context for global reasoning. Together, these innovations significantly enhance the representation, leading to better disparity initialization, as well as more powerful features for the subsequent iterative refinement process.  

Our contributions can be summarized as follows:
\begin{myitem}
    \item We present FoundationStereo, a zero-shot generalizable  stereo matching model that achieves comparable or even more favorable results to prior works fine-tuned on a target domain; it also significantly outperforms existing methods when applied to in-the-wild data.
    \item We create a large-scale (1M) high-fidelity synthetic dataset for stereo learning with high diversity and photorealism; and a self-curation pipeline to ensure that bad samples are pruned.
    \item To harness internet-scale knowledge containing rich semantic and geometric priors, we propose a Side-Tuning Adapter (STA) that adapts the ViT-based monocular depth estimation model~\cite{yang2024depthanythingv2} to the stereo setup.
    \item We develop Attentive Hybrid Cost Filtering (AHCF), which includes an hourglass module with 3D Axial-Planar Convolution (APC), and a Disparity Transformer (DT) module that performs full self-attention over the disparity dimension. 
\end{myitem}

\section{Related Work}

\boldparagraphstart{Deep Stereo Matching.} Recent advances in stereo matching have been driven by deep learning, significantly enhancing accuracy and generalization. Cost volume aggregation methods construct cost volumes from unary features and perform 3D CNN for volume filtering~\cite{xu2020aanet,yang2020waveletstereo,shen2021cfnet,mao2021uasnet,shen2022pcw,chen2023learning}, though the  high memory consumption prevents direct application to high resolution images. Iterative refinement methods, inspired by RAFT~\cite{teed2020raft}, bypasses the costly 4D volume construction and filtering by recurrently refining the disparity~\cite{lipson2021raft,zhao2023high,tosi2024neural,gong2024learning,crestereo,jing2023uncertainty,selective_stereo,chen2024mocha}. While they generalize well to various disparity range, the recurrent updates are often time-consuming, and lack long-range context reasoning. Recent works~\cite{igev,xu2024igevpp} thus combine the strengths of cost filtering and iterative refinement. With the tremendous progress made by vision transformers, another line of research~\cite{li2021revisiting,guo2022context,weinzaepfel2023croco} introduces transformer architecture to stereo matching, particularly in the unary feature extraction stage. Despite their success on per-domain fine-tuning setup, zero-shot generalization still remains challenging. To tackle this problem, \cite{zhang2020domain,zhang2025learning,chuah2022itsa,chang2023domain,rao2023masked,liu2022graftnet} explore learning domain-invariant features for cross-domain generalization, with a focus on training on Scene Flow~\cite{sceneflow2016} dataset. Concurrent work \cite{bartolomei2024stereo} achieves remarkable zero-shot generalization with monocular prior enhanced correlation volumes. However, the strong generalizability of vision foundation models emerged in other tasks that is supported by scaling law has yet to be fully realized in stereo matching for practical applications.

\input{table/dataset_comparison}

\boldparagraphstart{Stereo Matching Training Data.} Training data is essential for deep learning models. KITTI 12~\cite{geiger2012we} and KITTI 15~\cite{menze2015object} provide hundreds of training pairs on driving scenarios. DrivingStereo~\cite{yang2019drivingstereo} further scales up to 180K stereo pairs. Nevertheless, the sparse ground-truth disparity obtained by LiDAR sensors hinders learning accurate and dense stereo matching. Middlebury~\cite{middlebury} and ETH3D~\cite{eth3d} develop a low number of training data covering both indoor and outdoor scenarios beyond driving. Booster~\cite{ramirez2023booster} presents a real-world dataset focusing on transparent objects. InStereo2K~\cite{bao2020instereo2k} presents a larger training dataset consisting of 2K stereo pairs with denser ground-truth disparity obtained with structured light system. However, challenges of scarce data size, imperfect ground-truth disparity and lack of collection scalability in real-world have driven the widespread adoption of synthetic data for training. This includes Scene Flow~\cite{sceneflow2016}, Sintel~\cite{sintel}, CREStereo~\cite{crestereo}, IRS~\cite{wang2021irs}, TartanAir~\cite{tartanair}, FallingThings~\cite{fallingthings}, Virtual KITTI 2~\cite{cabon2020virtual}, CARLA HR-VS~\cite{yang2019hierarchical}, Dynamic Replica~\cite{karaev2023dynamicstereo}. In Tab.~\ref{tab:dataset_comparison}, we compare our proposed FoundationStereo dataset (FSD) with commonly used synthetic training datasets for stereo matching. Our dataset encompasses a wide range of scenarios, features the largest data volume to date, includes diverse 3D assets, captures stereo images under diversely randomized camera parameters, and achieves high fidelity in both rendering and spatial layouts.

\boldparagraphstart{Vision Foundation Models.}  Vision foundation models have significantly advanced across various vision tasks in 2D, 3D and multi-modal alignment. CLIP~\cite{radford2021learning} leverages large-scale image-text pair training to align visual and textual modalities, enabling zero-shot classification and facilitating cross-modal applications. DINO series~\cite{caron2021emerging,oquab2023dinov2,liu2023grounding} employ self-supervised learning for dense representation learning, effectively capturing detailed features critical for segmentation and recognition tasks.  SAM series~\cite{kirillov2023segment,ravi2024sam,yang2023track} demonstrate high versatility in segmentation driven by various prompts such as points, bounding boxes, language. Similar advancements also appear in 3D vision tasks. DUSt3R~\cite{wang2024dust3r} and MASt3R~\cite{leroy2024grounding} present generalizable frameworks for dense  3D reconstruction
from uncalibrated and unposed cameras. FoundationPose~\cite{wen2024foundationpose} develops a unified framework of 6D object pose estimation and tracking for novel objects. More closely related to this work, a number of efforts~\cite{yang2024depthanythingv2,yang2024depth,ke2024repurposing,bhat2023zoedepth} demonstrated strong generalization in monocular depth estimation task and multi-view stereo~\cite{izquierdo2025mvsanywhere}. 
Together, these approaches exemplify under the scaling law, how foundation models in vision are evolving to support robust applications across diverse scenarios without tedious per-domain fine-tuning.

\section{Approach}

\begin{figure*}[tbh]
    \centering
    {\includegraphics[width=0.95\textwidth]{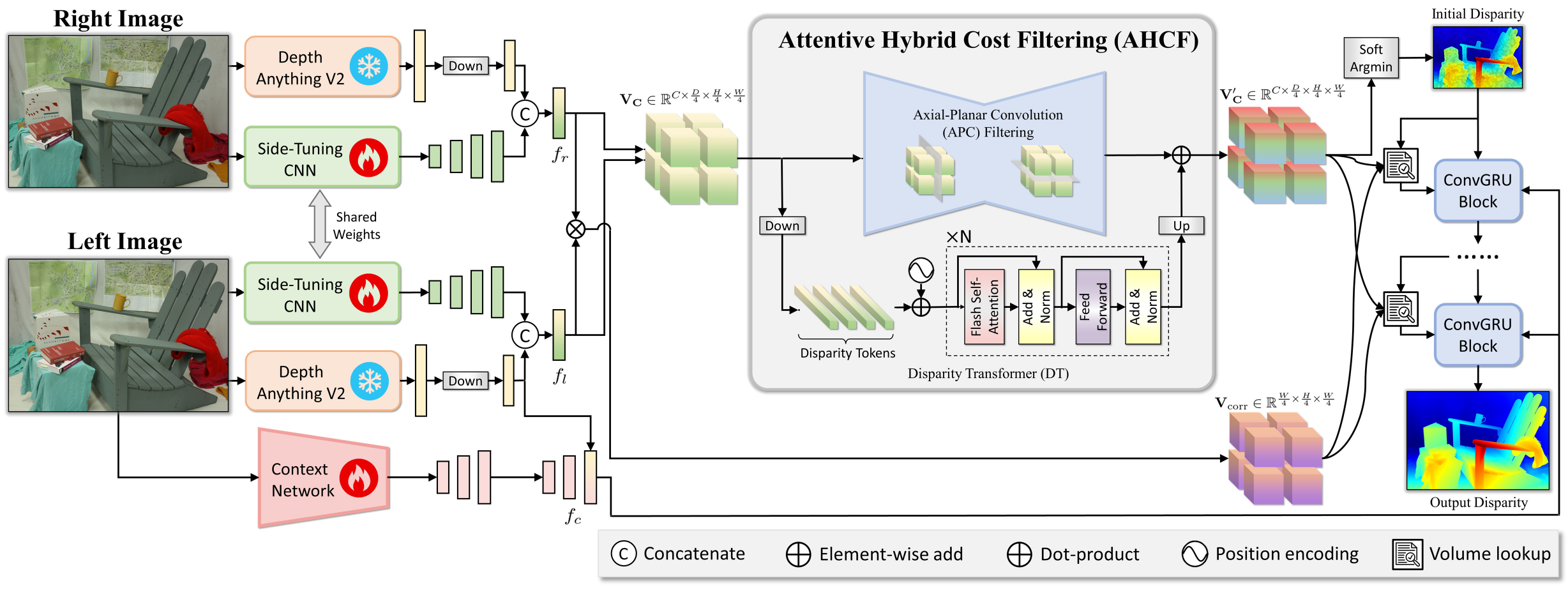}} 
    \vspace{-10pt}
    \caption{Overview of our proposed FoundationStereo. The Side-Tuning Adapter (STA) adapts the rich monocular priors from a frozen DepthAnythingV2~\cite{yang2024depthanythingv2}, while combined with fine-grained high-frequency features from multi-level CNN for unary feature extraction. Attentive Hybrid Cost Filtering (AHCF) combines the strengths of the Axial-Planar Convolution (APC) filtering and a Disparity Transformer (DT) module to effectively aggregate the features along spatial and disparity dimensions over the 4D hybrid cost volume. An initial disparity is then predicted from the filtered cost volume, and subsequently refined through GRU blocks. At each refinement step, the latest disparity is used to look up features from both filtered hybrid cost volume and correlation volume to guide the next refinement. The iteratively refined disparity becomes the final output.}  
    \label{fig:overview}
\end{figure*}

The overall network architecture is shown in Fig.~\ref{fig:overview}. 
The rest of this section describes the various components.

\subsection{Monocular Foundation Model Adaptation} \label{sec:side_tuning}

\begin{figure*}[h]
    \centering
    {\includegraphics[width=\textwidth]{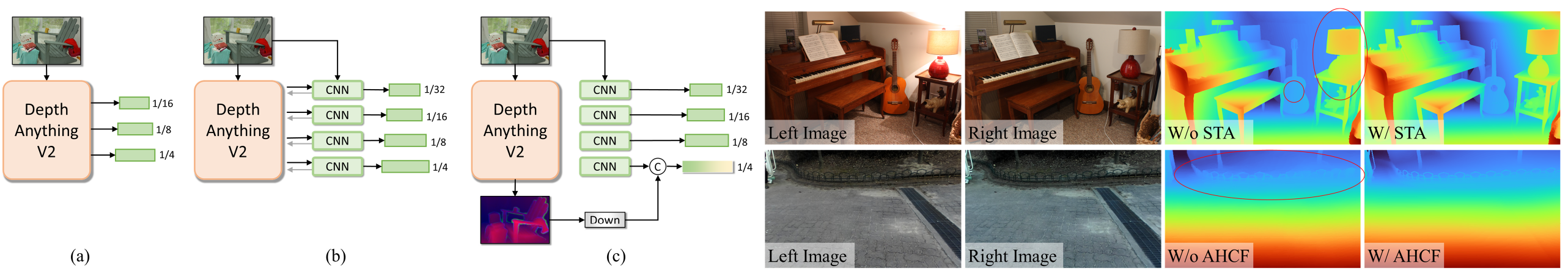}} 
    \vspace{-20pt}
    \caption{\textbf{Left:} Design choices for STA module.
    \textbf{Right:} Effects of the proposed STA and AHCF modules. ``W/o STA'' only uses CNN to extract features. ``W/o AHCF'' uses conventional 3D CNN-based hourglass network for cost volume filtering. Results are obtained via zero-shot inference without fine-tuning on target dataset. STA leverages rich monocular prior to reliably predict the lamp region with inconsistent lighting and dark guitar sound hole. AHCF effectively aggregates the spatial and long-range disparity context to accurately predict over thin repetitive structures.}  
    \label{fig:sta}
    \vspace{-10pt}
\end{figure*}

To mitigate the sim-to-real gap when the stereo network is primarily trained on synthetic dataset, we leverage the recent advancements on monocular depth estimation  trained on internet-scale real data~\cite{yang2024depthanythingv2,bochkovskii2024depth}. We use a CNN network to adapt the ViT-based monocular depth estimation network to the stereo setup, thus synergizing the strengths of both CNN and ViT architectures.

We explored multiple design choices for combining CNN and ViT approaches, as outlined in Fig.~\ref{fig:sta} (left). 
In particular, (a) directly uses the feature pyramids from the DPT head in a frozen DepthAnythingV2~\cite{yang2024depthanythingv2} without using CNN features. 
(b) resembles ViT-Adapter~\cite{vit_adapter} by exchanging features between CNN and ViT. (c) applies a $4\times 4$ convolution with stride 4 to downscale the feature before the DepthAnythingV2 final output head. The feature is then concatenated with the same level CNN feature to obtain a hybrid feature at 1/4 scale. The side CNN network is thus learned to adapt the ViT features~\cite{zhang2020side} to stereo matching task. Surprisingly, while being simple, we found (c) significantly surpasses the alternative choices on the stereo matching task, as shown in the experiments (Sec.~\ref{sec:ablation}). 
As a result, we adopt (c) as the main design of STA module. 

Formally, given a pair of  left and right images $I_l, I_r\in \mathbb{R}^{H\times W \times 3}$, we employ EdgeNeXt-S~\cite{maaz2022edgenext} as the CNN module within STA to extract multi-level pyramid features, where the 1/4 level feature is equipped with DepthAnythingV2 feature: $ f_l^{(i)}, f_r^{(i)} \in \mathbb{R}^{C_i \times \frac{H}{i} \times \frac{W}{i}} $, $i \in \{4, 8, 16, 32\}$. EdgeNeXt-S~\cite{maaz2022edgenext} is chosen for its memory efficiency and because larger CNN backbones did not yield additional benefits in our investigation. When forwarding to DepthAnythingV2, we first resize the image to be divisible by 14, to be consistent with its pretrained patch size.  The STA weights are shared when applied to $I_l, I_r$. 

Similarly, we employ STA to extract context feature, with the difference that the CNN module is designed with a sequence of residual blocks~\cite{he2016deep} and down-sampling layers. It  generates context features of multiple scales: $f_c^{(i)}  \in \mathbb{R}^{C_i \times \frac{H}{i} \times \frac{W}{i}} $, $i \in \{4, 8, 16\}$, as in \cite{lipson2021raft}.  $f_c$ participates in  initializing the hidden state of the ConvGRU block and inputting to the ConvGRU block at each iteration, effectively guiding the iterative process with progressively refined contextual information. 

Fig.~\ref{fig:sta} visualizes the power of rich monocular prior that helps to reliably predict on ambiguous regions which is challenging to deal with by naive correspondence search along the epipolar line. Instead of using the raw monocular depth from DepthAnythingV2 which has scale ambiguity, we use its latent feature as geometric priors extracted from both stereo images and compared through cost filtering as described next.

\subsection{Attentive Hybrid Cost Filtering} \label{sec:cost_volume}

\boldparagraphstart{Hybrid Cost Volume Construction.} Given unary features at 1/4 scale $f_l^{4},f_r^{4}$ extracted from previous step, we construct the cost volume $\mathbf{V_C} \in \mathbb{R}^{C\times \frac{D}{4} \times \frac{H}{4} \times \frac{W}{4}}$ with a combination of group-wise correlation and concatenation~\cite{guo2019group}:
\begin{align}
   &\mathbf{V_{\text{gwc}}}(g,d,h,w) = \left \langle \widehat{f}_{l,g}^{(4)}(h,w),  
    \widehat{f}_{r,g}^{(4)}(h,w-d) \right \rangle, \notag \\
    &\mathbf{V_{\text{cat}}}(d,h,w) = \left[ \text{Conv}(f_{l}^{(4)})(h,w), \text{Conv}(f_{r}^{(4)})(h,w-d) \right], \notag \\
    &\mathbf{V_C}(d,h,w) = \left[ \mathbf{V_{\text{gwc}}}(d,h,w), \mathbf{V_{\text{cat}}}(d,h,w) \right]  \label{eq:V_C}
\end{align}
where $\widehat{f}$ denotes $L_2$ normalized feature for better training stability; $\left\langle \cdot, \cdot \right\rangle$ represents dot product; $g\in \{1,2,...,G\}$ is the group index among the total $G=8$ feature groups that we evenly divide the total features into; $d\in \{1,2,...,\frac{D}{4}\}$ is the disparity index. $[\cdot, \cdot]$ denotes concatenation along channel dimension. The group-wise correlation $\mathbf{V_{\text{gwc}}}$ harnesses the strengths of conventional correlation-based matching costs, offering a diverse set of similarity measurement features from each group. $\mathbf{V_{\text{cat}}}$ preserves unary features  including the rich monocular priors by concatenating left and right features at shifted disparity. To reduce memory consumption, we linearly downsize the unary feature dimension to 14 using a convolution of kernel size 1 (weights are shared between $f_{l}^4$ and $f_{r}^4$) before concatenation. Next, we describe two sub-modules for effective cost volume filtering.

\boldparagraphstart{Axial-Planar Convolution (APC) Filtering.} An hourglass network consisting of 3D convolutions, with three down-sampling blocks and three up-sampling blocks with residual connections, is leveraged for cost volume filtering~\cite{igev,bangunharcana2021correlate}. While 3D convolutions of kernel size $3\times 3 \times 3$ are commonly used for relatively small disparity sizes~\cite{igev,guo2019group,chang2018pyramid}, we observe it struggles with larger disparities  when applied to high resolution images, especially since the disparity dimension is expected to model the probability distribution for the initial disparity prediction. However, it is impractical to naively increase the kernel size, due to the intensive memory consumption. In fact, even when setting kernel size to $5\times 5\times 5$ we observe unmanageable memory usage on an 80 GB GPU. This drastically limits the model's representation power when scaling up with large amount of training data.  We thus develop ``Axial-Planar Convolution'' which decouples a single $3\times 3 \times 3$  convolution into  two separate convolutions:  one over spatial dimensions (kernel size $K_s\times K_s\times 1$) and the other over disparity ($1\times 1\times K_d$), each followed by BatchNorm and ReLU. 
APC can be regarded as a 3D version of Separable Convolution~\cite{chollet2017xception} with the difference that we only separate the spatial and disparity dimensions without subdividing the channel into groups which sacrifices representation power. The disparity dimension is specially treated due to its uniquely encoded feature comparison within the cost volume. 
We use APC wherever possible in the hourglass network except for the down-sampling and up-sampling layers.

\boldparagraphstart{Disparity Transformer (DT).} While prior works~\cite{weinzaepfel2023croco,li2021revisiting} introduced transformer architecture to unary feature extraction step to scale up stereo training, the cost filtering process is often overlooked, which remains an essential step in achieving accurate stereo matching by  encapsulating correspondence information. Therefore, we introduce DT to further enhance the long-range context reasoning within the 4D cost volume. Given $\mathbf{V_C}$ obtained in Eq.~\eqref{eq:V_C}, we first apply a 3D convolution of kernel size $4\times 4\times 4$ with stride 4 to downsize the cost volume. We then reshape the volume into a batch of token sequences, each with length of disparity. We apply position encoding before feeding it to a series (4 in our case) of transformer encoder blocks, where FlashAttention~\cite{dao2022flashattention} is leveraged to perform multi-head self-attention~\cite{vaswani2017attention}. The process can be written as:
\begin{align}
    &\mathbf{Q_0}=\text{PE}\left( \mathbf{R}\left(\text{Conv}_{4\times 4\times 4}(\mathbf{V_C}) \right) \right) \in \mathbb{R}^{\left(\frac{H}{16}\times \frac{W}{16}\right) \times C\times \frac{D}{16}} \notag \\
    &\text{MultiHead}(\mathbf{Q}, \mathbf{K}, \mathbf{V}) = [\text{head}_1, \ldots, \text{head}_h]\mathbf{W}_O \notag \\
    &\quad \text{where } \text{head}_i = \text{FlashAttention}(\mathbf{Q}_i, \mathbf{K}_i, \mathbf{V}_i) \notag \\
    &\mathbf{Q_1}=\text{Norm}\left( \text{MultiHead}(\mathbf{Q_0}, \mathbf{Q_0}, \mathbf{Q_0}) + \mathbf{Q_0} \right) \notag \\
    &\mathbf{Q_2}=\text{Norm}\left( \text{FFN}(\mathbf{Q_1}) + \mathbf{Q_1} \right) \notag 
\end{align} 
where $\mathbf{R}(\cdot)$ denotes reshape operation; $\text{PE}(\cdot)$ represents position encoding; $[\cdot,\cdot]$ denotes concatenation along the channel dimension; $\mathbf{W}_O$ is linear weights. The number of heads is $h=4$ in our case. Finally, the DT output is up-sampled to the same size as $\mathbf{V_C}$ using trilinear interpolation and summed with hourglass output, as shown in Fig.~\ref{fig:overview}.

\boldparagraphstart{Initial Disparity Prediction.} We apply soft-argmin~\cite{kendall2017end} to the filtered volume $\mathbf{V_C'}$ to produce an initial disparity:
\begin{align}
    d_0=\sum_{d=0}^{\frac{D}{4}-1}d\cdot \text{Softmax}(\mathbf{V_C'})(d)
\end{align}
where $d_0$ is at 1/4 scale of the original image resolution.

\vspace{-4pt}
\subsection{Iterative Refinement}\label{sec:refine}
Given $d_0$, we perform iterative GRU updates to progressively refine disparity, which helps to avoid local optimum and accelerate convergence~\cite{igev}. In general, the k-th update can be formulated as:
\begin{align}
    &\mathbf{V}_{\text{corr}}(w',h,w)=\left\langle f_l^{(4)}(h,w), f_r^{(4)}(h,w') \right\rangle  \\
    &\mathbf{F_V}(h,w)=[\mathbf{V_C'}(d_k,h,w), \mathbf{V}_{\text{corr}}(w-d_k,h,w)] \\
    &x_k=[\text{Conv}_v(\mathbf{F_V}), \text{Conv}_{d}(d_k), d_k, c]  \\
    &z_k=\sigma\left( \text{Conv}_z([h_{k-1}, x_k]) \right) \\
    &r_k=\sigma\left( \text{Conv}_r([h_{k-1}, x_k])  \right)  \\
    &\hat{h}_k=\text{tanh}\left( \text{Conv}_h([r_k \odot h_{k-1}, x_k]) \right) \\
    &h_k=(1-z_k)\odot h_{k-1} + z_k \odot \hat{h}_k  \\
    &d_{k+1} = d_k + \text{Conv}_{\Delta}(h_k) 
\end{align}
where  $\odot$ denotes element-wise product; $\sigma$ denotes sigmoid; $\mathbf{V}_{\text{corr}} \in \mathbb{R}^{\frac{W}{4}\times \frac{H}{4} \times \frac{W}{4}}$ is the pair-wise correlation volume; $\mathbf{F_V}$ represents the looked up volume features using latest disparity; $c=\text{ReLU}(f_c)$ encodes the context feature from left image, including STA adapted features (Sec.~\ref{sec:side_tuning}) which effectively guide the refinement process leveraging rich monocular priors.

\begin{figure*}[h]
    \centering
    \vspace{-10pt}
    {\includegraphics[width=0.98\textwidth]{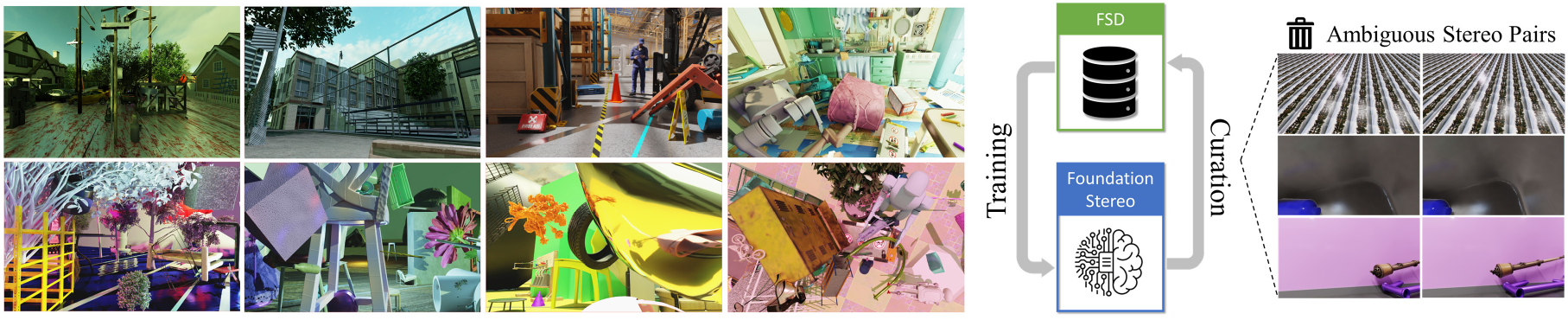}} 
    \vspace{-10pt}
    \caption{\textbf{Left:} Samples from our FoundationStereo dataset (FSD), which consists of synthetic stereo images with structured indoor / outdoor scenes (top), as well as more randomized scenes with challenging flying objects and higher geometry and texture diversity (bottom). \textbf{Right:} The iterative self-curation process removes ambiguous samples inevitably produced from the domain randomized synthetic data generation process. Example ambiguities include severe texture repetition, ubiquitous reflections with limited surrounding context, and pure color under improper lighting.}  
    \label{fig:sdg_montage}
    \vspace{-10pt}
\end{figure*}

We use three levels of GRU blocks to perform coarse-to-fine hidden state update in each iteration, where the initial hidden states are produced from context features $h_0^{(i)}=\text{tanh}(f_c^{(i)}), i \in \{4,8,16\}$. At each level, attention-based selection mechanism~\cite{selective_stereo} is leveraged to capture information at different frequencies. Finally, $d_k$ is up-sampled to the full resolution using convex sampling~\cite{teed2020raft}.

\subsection{Loss Function}
\label{sec:costfn}

The model is trained with the following objective:
\begin{align}
    \mathcal{L}=\left| d_0-\overline{d}  \right|_{\text{smooth}} + \sum_{k=1}^{K}\gamma^{K-k}\left\| d_k-\overline{d} \right\|_1
\end{align}
where $\overline{d}$ represents ground-truth disparity; $\left|\cdot \right|_{\text{smooth}}$ denotes smooth $L_1$ loss;  $k$ is the iteration number; $\gamma$ is set to 0.9, and we apply exponentially increasing weights~\cite{lipson2021raft} to supervise the iteratively refined disparity.

\subsection{Synthetic Training Dataset}\label{sec:sdg}

We created a large scale synthetic training dataset with NVIDIA Omniverse.  This FoundationStereo Dataset (FSD) accounts for crucial stereo matching challenges such as reflections, low-texture surfaces, and severe occlusions. We perform domain randomization~\cite{tobin2017domain} to augment dataset diversity, including random stereo baseline, focal length, camera perspectives, lighting conditions and object configurations. Meanwhile, high-quality 3D assets with abundant textures and path-tracing rendering are leveraged to enhance realism in rendering and layouts. Fig.~\ref{fig:sdg_montage} displays some samples from our dataset including both structured indoor and outdoor scenarios, as well as more diversely randomized flying objects with various geometries and textures under complex yet realistic lighting. 
See the appendix for details.

\boldparagraphstart{Iterative Self-Curation.} While synthetic  data generation in theory can produce unlimited amount of data and achieve large diversity through randomization, ambiguities can  be inevitably introduced especially for less structured scenes with flying objects, which confuses the learning process. To eliminate those samples, we design an automatic  iterative self-curation strategy. Fig.~\ref{fig:sdg_montage} demonstrates this process and detected ambiguous samples. We start with training an initial version of FoundationStereo on FSD, after which it is  evaluated on  FSD. Samples where BP-2 (Sec.~\ref{sec:datasets}) is larger than 60\% are regarded as ambiguous samples and replaced by regenerating new ones. The training and curation processes are alternated to iteratively (twice in our case) update both FSD and FoundationStereo.

\section{Experiments}

\subsection{Implementation Details}\label{sec:implementation}

We implement FoundationStereo in PyTorch. The foundation model is trained on a mixed dataset consisting of our proposed FSD, together with Scene Flow~\cite{sceneflow2016}, Sintel~\cite{sintel}, CREStereo~\cite{crestereo}, FallingThings~\cite{fallingthings}, InStereo2K~\cite{bao2020instereo2k} and Virtual KITTI~2~\cite{cabon2020virtual}. We train FoundationStereo using AdamW optimizer~\cite{loshchilov2017decoupled} for 200K steps with a total batch size of 128 evenly distributed over 32 NVIDIA A100 GPUs. The learning rate starts at 1e-4 and decays by 0.1 at 0.8 of the entire training process. Images are randomly cropped to 320$\times$736 before feeding to the network. Data augmentations similar to \cite{lipson2021raft} are performed. During training, 22 iterations are used in GRU updates. In the following, unless otherwise mentioned, we use the same foundation model for zero-shot inference with 32 refinement iterations and 416 for maximum disparity. 

\subsection{Benchmark Datasets and Metric}\label{sec:datasets}
\vspace{-5pt}
\boldparagraphstart{Datasets.} We consider five commonly used public datasets for evaluation: Scene Flow~\cite{sceneflow2016} is a synthetic dataset including three subsets: FlyingThings3D, Driving, and Monkaa. 
Middlebury~\cite{middlebury} consists of indoor stereo image pairs with high-quality ground-truth disparity captured via structured light. Unless otherwise mentioned, evaluations are performed on half resolution and non-occluded regions.
ETH3D~\cite{eth3d} provides grayscale stereo image pairs covering both indoor and outdoor scenarios. 
KITTI 2012~\cite{geiger2012we} and KITTI 2015~\cite{menze2015object} datasets feature real-world driving scenes, where sparse ground-truth disparity maps are provided, which are derived from LIDAR sensors.

\boldparagraphstart{Metrics.} ``EPE'' computes average per-pixel disparity error. ``BP-X'' computes the percentage of pixels where the disparity error is larger than X pixels. ``D1''  computes the percentage of pixels whose disparity error is larger than 3 pixels and 5\% of the ground-truth disparity.

\subsection{Zero-Shot Generalization Comparison}

\input{table/zero_shot}

\boldparagraphstart{Benchmark Evaluation.} Tab.~\ref{tab:zero_shot} exhibits quantitative comparison of zero-shot generalization results on four public real-world datasets. Even when trained solely on Scene Flow, our method outperforms the comparison methods consistently across all datasets, thanks to the efficacy of adapting rich monocular priors from vision foundation models. We further evaluate in a more realistic setup, allowing methods to train on any available dataset while excluding the target domain, to achieve optimal zero-shot inference results as required in practical applications.

\begin{figure*}[tbh]
    \centering
    {\includegraphics[width=0.99\textwidth]{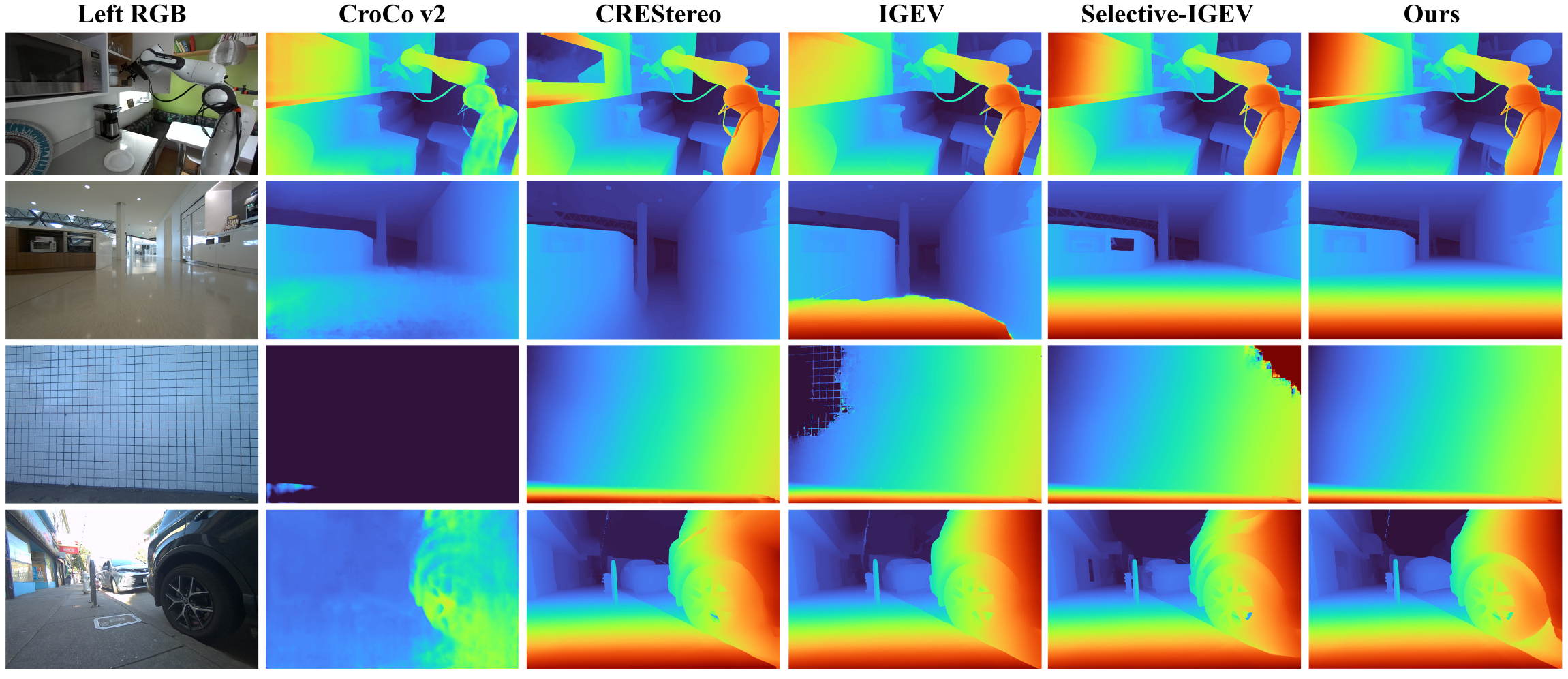}} 
    \vspace{-10pt}
    \caption{Qualitative comparison of zero-shot inference on in-the-wild images. For each comparison method we select the best performing checkpoint from their public release, which has been trained on a mixture of public datasets. These images exhibit challenging reflection, translucency, repetitive textures, complex illuminations and thin-structures, revealing the importance of our network architecture and large-scale training.}  
    \label{fig:wild}
    \vspace{-5pt}
\end{figure*}

\input{table/sceneflow}

\input{table/eth_finetune}

\boldparagraphstart{In-the-Wild Generalization.} We compare our foundation model against recent approaches that released their checkpoints trained on a mixture of datasets, to resemble the practical zero-shot application on in-the-wild images. Comparison methods   include CroCo v2~\cite{weinzaepfel2023croco}, CREStereo~\cite{crestereo}, IGEV~\cite{igev} and Selective-IGEV~\cite{selective_stereo}. For each method, we select the best performing checkpoint from their public release. In this evaluation, the four real-world benchmark datasets~\cite{middlebury,eth3d,geiger2012we,menze2015object} have been used for training comparison methods, whereas they are not used in our fixed foundation model.  Fig.~\ref{fig:wild} displays qualitative comparison on various scenarios, including a robot scene from DROID~\cite{khazatsky2024droid} dataset and custom captures covering indoor and outdoor.

\subsection{In-Domain Comparison}

Tab.~\ref{tab:sceneflow} presents quantitative comparison on Scene Flow, where all methods are following the same officially divided train and test split. Our FoundationStereo model outperforms the comparison methods by a large margin, reducing the previous best EPE from 0.41 to 0.33.  Although in-domain training is not the focus of this work, the results reflect the effectiveness of our model design.

Tab.~\ref{tab:eth_finetune} exhibits quantitative comparison on ETH3D leaderboard (test set). For our approach, we perform evaluations in two settings. First, we fine-tune our foundation model on a mixture of the default training dataset (Sec.~\ref{sec:implementation})  and ETH3D training set for another 50K steps, using the same learning rate schedule and data augmentation. Our model significantly surpasses the previous best approach by reducing more than half of the error rates and ranks 1st on leaderboard at the time of submission. 
This indicates great potential of transferring capability from our foundation model if in-domain fine-tuning is desired. Second, we also evaluated our foundation model without using any data from ETH3D. Remarkably, our foundation model's  zero-shot inference achieves comparable or even better results than leading approaches that perform in-domain training.

In addition, our finetuned model also ranks 1st on the Middlebury leaderboard. See appendix for details.

\subsection{Ablation Study}\label{sec:ablation}

\input{table/ablation_foundation}

\input{table/ablation_ahcf}

We investigate different design choices for our model and dataset. Unless otherwise mentioned, we train on a randomly subsampled version (100K) of FSD to make the experiment scale more affordable. Given Middlebury dataset's high quality ground-truth, results are  evaluated on its training set to reflect zero-shot generalization. Since the focus of this work is to build a stereo matching foundation model with strong generalization, we  do not deliberately limit model size while pursuing better performance.

\boldparagraphstart{STA Design Choices.} As shown in Tab.~\ref{tab:sta_ablation}, we first compare different vision foundation models for adapting rich monocular priors, including different model sizes of DepthAnythingV2~\cite{yang2024depthanythingv2} and DINOv2-Large~\cite{oquab2023dinov2}. While DINOv2  previously exhibited promising results in  correspondence matching~\cite{el2024probing}, it is not as effective as DepthAnythingV2 in the stereo matching task, possibly due to its less task-relevance and its limited resolution to reason high-precision pixel-level correspondence. We then study different design choices from Fig.~\ref{fig:sta}.  Surprisingly,
while being simple, we found (c) significantly surpasses the alternatives. We hypothesize the latest feature
before the final output head preserves high-resolution and fine-grained semantic and geometric priors that are suitable for subsequent cost volume construction and filtering process. We also experimented whether to freeze the adapted ViT model. As expected, unfreezing ViT corrupts the pretrained monocular priors, leading to degraded performance. 

\input{table/ablation_overall}

\boldparagraphstart{AHCF Design Choices.} As shown in Tab.~\ref{tab:ablation_ahcf}, for DT module we study different position embedding (row 1-2); different feature scale to perform transformer (row 3-4); transformer over the full cost-volume or only along the disparity dimension (row 5-6); different placements of DT module relative to the hourglass network (row 7-9). Specifically, RoPE~\cite{su2024roformer} encodes relative distances between tokens instead of absolute positions, making it more adaptive to varying sequence lengths. However, it does not outperform cosine position embedding, probably due to the constant disparity size in 4D cost volume. While in theory, full volume attention provides larger receptive field, it is less effective than merely applying over the disparity dimension of the cost volume. We hypothesize the extremely large space of 4D cost volume makes it less tractable, whereas attention over disparity provides sufficient context for a better initial disparity prediction and subsequent volume feature lookup during GRU updates. Next, we compare different kernel sizes in APC (row 10-15), where the last dimension in each parenthesis corresponds to disparity dimension. We observe increasing benefits when enlarging disparity kernel size until it saturates at around 17.

\boldparagraphstart{Effects of Proposed Modules.} The quantitative effects are shown in Tab.~\ref{tab:ablation_overall} (left). STA leverages rich monocular priors which greatly enhances generalization to real images for ambiguous regions. DT and APC effectively aggregate cost volume features along spatial and disparity dimensions, leading to improved context for disparity initialization and subsequent volume feature look up during GRU updates. Fig.~\ref{fig:sta} further visualizes the resulting effects.

\boldparagraphstart{Effects of FoundationStereo Dataset.} We study whether to include FSD dataset with the existing public datasets for training our foundation model described in Sec.~\ref{sec:implementation}. Results are shown  in Tab.~\ref{tab:ablation_overall} (right).

\section{Conclusion}
We introduced FoundationStereo, a foundation model for stereo depth estimation that achieves strong zero-shot generalization across various domains without fine-tuning. We envision such a foundation model can facilitate broader adoption of stereo estimation models in practical applications. Despite its remarkable generalization, it has several limitations. First, our model is not yet optimized for efficiency, which takes 0.7s on image size of 375$\times$1242 on NVIDIA A100 GPU. Future work could explore adapting distillation and pruning techniques applied to other vision foundation models~\cite{zhao2023fast,chen20230}. Second, our dataset FSD includes a limited collection of transparent objects. Robustness could be further enhanced by augmenting with a larger diversity of fully transparent objects during training.

{
    \small
    \bibliographystyle{ieeenat_fullname}
    \bibliography{ref}
}

\input{suppl}

\end{document}

%% file: table/dataset_comparison.tex
\begin{table*}[tbh]
\centering
\def\mywidth{0.95\textwidth} 
\definecolor{green}{RGB}{0,200,0}
\resizebox{\mywidth}{!}{

\begin{tabular}{crccccccccc}
\thickline
\multicolumn{2}{c}{Properties}                                  & Sintel~\cite{sintel}           & Sceneflow~\cite{sceneflow2016} & CREStereo~\cite{crestereo}     & IRS~\cite{wang2021irs}         & TartanAir~\cite{tartanair}     & FallingThings~\cite{fallingthings} & UnrealStereo4K~\cite{tosi2021smd} & Spring~\cite{Mehl2023_Spring}  & FSD (Ours) \bigstrut\\
\hline
\multirow{5}[1]{*}{\begin{sideways}Scenarios\end{sideways}} & \multicolumn{1}{l}{Flying Objects} & \redxmark                      & \greencheckmark                & \greencheckmark                & \redxmark                      & \redxmark                      & \redxmark                      & \redxmark                      & \redxmark                      & \greencheckmark \bigstrut[t]\\
                               & \multicolumn{1}{l}{Indoor}     & \redxmark                      & \redxmark                      & \redxmark                      & \greencheckmark                & \greencheckmark                & \greencheckmark                & \greencheckmark                & \redxmark                      & \greencheckmark \\
                               & \multicolumn{1}{l}{Outdoor}    & \redxmark                      & \greencheckmark                & \redxmark                      & \redxmark                      & \greencheckmark                & \greencheckmark                & \greencheckmark                & \redxmark                      & \greencheckmark \\
                               & \multicolumn{1}{l}{Driving}    & \redxmark                      & \greencheckmark                & \redxmark                      & \redxmark                      & \redxmark                      & \redxmark                      & \redxmark                      & \redxmark                      & \greencheckmark \\
                               & \multicolumn{1}{l}{Movie}      & \greencheckmark                & \greencheckmark                & \redxmark                      & \redxmark                      & \redxmark                      & \redxmark                      & \redxmark                      & \greencheckmark                & \redxmark \\
\rowcolor[rgb]{ .851,  .851,  .851} \multicolumn{2}{c}{Simulator}                                   & Blender                        & Blender                        & Blender                        & Unreal Engine                  & Unreal Engine                  & Unreal Engine                  & Unreal Engine                  & Blender                        & NVIDIA Omniverse \\
\multicolumn{2}{c}{Rendering Realism}                           & \textcolor[rgb]{ 0,  .498,  0}{High} & \textcolor[rgb]{ .753,  0,  0}{Low} & \textcolor[rgb]{ 0,  .498,  0}{High} & \textcolor[rgb]{ 0,  .498,  0}{High} & \textcolor[rgb]{ 0,  .498,  0}{High} & \textcolor[rgb]{ 0,  .498,  0}{High} & \textcolor[rgb]{ 0,  .498,  0}{High} & \textcolor[rgb]{ 0,  .498,  0}{High} & \textcolor[rgb]{ 0,  .498,  0}{High} \\
\rowcolor[rgb]{ .847,  .847,  .847} \multicolumn{2}{c}{Scenes}                                      & 10                             & 9                              & 0                              & 4                              & 18                             & 3                              & \cellcolor[rgb]{ .851,  .851,  .851}8 & \cellcolor[rgb]{ .851,  .851,  .851}47 & 12 \\
\multicolumn{2}{c}{Layout Realism}                              & \textcolor[rgb]{ .749,  .565,  0}{Medium} & \textcolor[rgb]{ .753,  0,  0}{Low} & \textcolor[rgb]{ .753,  0,  0}{Low} & \textcolor[rgb]{ 0,  .498,  0}{High} & \textcolor[rgb]{ 0,  .498,  0}{High} & \textcolor[rgb]{ 1,  .753,  0}{Medium} & \textcolor[rgb]{ 0,  .498,  0}{High} & \textcolor[rgb]{ 0,  .498,  0}{High} & \textcolor[rgb]{ 0,  .498,  0}{High} \\
\rowcolor[rgb]{ .847,  .847,  .847} \multicolumn{2}{c}{Stereo Pairs}                                & \cellcolor[rgb]{ .851,  .851,  .851}1K\textsuperscript{\dag} & \cellcolor[rgb]{ .851,  .851,  .851}40K\textsuperscript{\dag} & \cellcolor[rgb]{ .851,  .851,  .851}200K & \cellcolor[rgb]{ .851,  .851,  .851}103K\textsuperscript{\dag} & \cellcolor[rgb]{ .851,  .851,  .851}306K\textsuperscript{\dag} & \cellcolor[rgb]{ .851,  .851,  .851}62K & \cellcolor[rgb]{ .851,  .851,  .851}7.7K & \cellcolor[rgb]{ .851,  .851,  .851}6K\textsuperscript{\dag} & 1000K \\
\multicolumn{2}{c}{Resolution}                                  & $1024\times 436$               & $960\times 540$                & $1920\times 1080$              & $960 \times 540$               & $640\times 480$                & $960\times 540$                & $3840\times 2160$              & $1920\times 1080$              & $1280\times 720$ \\
\rowcolor[rgb]{ .847,  .847,  .847} \multicolumn{2}{c}{Reflections}                                 & \redxmark                      & \redxmark                      & \greencheckmark                & \greencheckmark                & \greencheckmark                & \greencheckmark                & \greencheckmark                & \redxmark                      & \greencheckmark \\
\multicolumn{2}{c}{Camera Params}                               & \textcolor[rgb]{ .753,  0,  0}{Constant} & \textcolor[rgb]{ .753,  0,  0}{Constant} & \textcolor[rgb]{ .753,  0,  0}{Constant} & \textcolor[rgb]{ .753,  0,  0}{Constant} & \textcolor[rgb]{ .753,  0,  0}{Constant} & \textcolor[rgb]{ .753,  0,  0}{Constant} & \textcolor[rgb]{ .753,  0,  0}{Constant} & \textcolor[rgb]{ 1,  .753,  0}{\makecell{Constant baseline,\\varying intrinsics}} & \textcolor[rgb]{ 0,  .498,  0}{\makecell{Varying baseline\\ and intrinsics}} \\
\thickline
\end{tabular}%

}
\vspace{-10pt}
\caption{Synthetic datasets for training stereo algorithms (excluding test images with inaccessible ground truth). 
$^\dag$Indicates reduced diversity, caused by including many similar frames from video sequences.}
\vspace{-10pt}
\label{tab:dataset_comparison}
\end{table*}

%% file: table/zero_shot.tex
\begin{table}[t]
\centering
\def\mywidth{0.49\textwidth} 
\definecolor{green}{RGB}{0,200,0}
\resizebox{\mywidth}{!}{

\begin{tabular}{lrrrr}
\hline
\multirow{2}[2]{*}{Methods}    & Middlebury                     &  ETH3D                         & KITTI-12                       & KITTI-15  \bigstrut[t]\\
                               & BP-2                           & BP-1                           & D1                             & D1 \bigstrut[b]\\
\hline
CREStereo++~\cite{jing2023uncertainty} & 14.8                           & 4.4                            & 4.7                            & 5.2 \bigstrut[t]\\
DSMNet~\cite{zhang2020domain}  & 13.8                           & 6.2                            & 6.2                            & 6.5 \\
Mask-CFNet~\cite{rao2023masked} & 13.7                           & 5.7                            & 4.8                            & 5.8 \\
HVT-RAFT~\cite{chang2023domain} & 10.4                           & 3.0                            & 3.7                            & 5.2 \\
RAFT-Stereo~\cite{lipson2021raft} & 12.6                            & 3.3                            & 4.7                            & 5.5 \\
Selective-IGEV~\cite{selective_stereo} & 9.2                            & 5.7                            & 4.5                            & 5.6 \\
IGEV~\cite{lipson2021raft}     & 8.8                            & 4.0                            & 5.2                            & 5.7 \\
Former-RAFT-DAM~\cite{zhang2025learning} & 8.1                            & 3.3                            & 3.9                            & 5.1 \\
IGEV++~\cite{xu2024igevpp}     & 7.8                            & 4.1                            & 5.1                            & 5.9 \\
NMRF~\cite{guan2024neural}     & 7.5                            & 3.8                            & 4.2                            & 5.1 \\
\rowcolor[rgb]{ .886,  .937,  .851} Ours (Scene Flow)              & \textbf{5.5}                   & \textbf{1.8}                   & \textbf{3.2}                   & \textbf{4.9} \bigstrut[b]\\
\hline
Selective-IGEV*~\cite{selective_stereo} & 7.5                            & 3.4                            & 3.2                            & 4.5 \bigstrut[t]\\
\rowcolor[rgb]{ .886,  .937,  .851} Ours                           & \textbf{1.1}                   & \textbf{0.5}                   & \textbf{2.3}                   & \textbf{2.8} \bigstrut[b]\\
\hline
\end{tabular}%

}
\vspace{-8pt}
\caption{Zero-shot generalization results on four public datasets. The most commonly used metrics for each dataset were adopted. In the first block, all methods were trained only on Scene Flow. In the second block, methods are allowed to train on any existing datasets excluding the four target domains. The weights and parameters are fixed for evaluation.}
\label{tab:zero_shot}
\vspace{-15pt}
\end{table}

%% file: table/sceneflow.tex
\begin{table*}[h]
\centering
\def\mywidth{0.95\textwidth} 
\definecolor{green}{RGB}{0,200,0}
\resizebox{\mywidth}{!}{

\begin{tabular}{ccccccccc}
\toprule
Method                         & \multicolumn{1}{l}{LEAStereo~\cite{cheng2020hierarchical}} & \multicolumn{1}{l}{GANet~\cite{zhang2019ga}} & \multicolumn{1}{l}{ACVNet~\cite{xu2022attention}} & \multicolumn{1}{l}{IGEV-Stereo~\cite{igev}} & NMRF~\cite{guan2024neural}     & MoCha-Stereo~\cite{chen2024mocha} & \multicolumn{1}{l}{Selective-IGEV~\cite{selective_stereo}} & \cellcolor[rgb]{ .886,  .937,  .855}Ours \bigstrut[t]\\
EPE                            & 0.78                           & 0.84                           & 0.48                           & 0.47                           & 0.45                           & 0.41                           & 0.44                           & \cellcolor[rgb]{ .886,  .937,  .855}\textbf{0.34} \bigstrut[b]\\
\bottomrule
\end{tabular}%

}
\vspace{-8pt}
\caption{Comparison of methods trained / tested on the Scene Flow train / test sets, respectively.} \label{tab:sceneflow}
\vspace{-10pt}
\end{table*}

%% file: table/eth_finetune.tex
\begin{table}[tbh]
\centering
\def\mywidth{0.4\textwidth} 
\definecolor{green}{RGB}{0,200,0}
\resizebox{\mywidth}{!}{

\begin{tabular}{lcccc}
\toprule
Method                         & Zero-Shot                      & BP-0.5                         & BP-1.0                         & EPE \\
\midrule
GMStereo~\cite{gmstereo}       & \redxmark                      & 5.94                           & 1.83                           & 0.19 \\
HITNet~\cite{tankovich2021hitnet} & \redxmark                      & 7.83                           & 2.79                           & 0.20 \\
EAI-Stereo~\cite{zhao2022eai}  & \redxmark                      & 5.21                           & 2.31                           & 0.21 \\
RAFT-Stereo~\cite{lipson2021raft} & \redxmark                      & 7.04                           & 2.44                           & 0.18 \\
CREStereo~\cite{crestereo}     & \redxmark                      & 3.58                           & 0.98                           & 0.13 \\
IGEV-Stereo~\cite{igev}        & \redxmark                      & 3.52                           & 1.12                           & 0.14 \\
CroCo-Stereo~\cite{weinzaepfel2023croco} & \redxmark                      & 3.27                           & 0.99                           & 0.14 \\
MoCha-Stereo~\cite{chen2024mocha} & \redxmark                      & 3.20                           & 1.41                           & 0.13 \\
Selective-IGEV~\cite{selective_stereo} & \redxmark                      & 3.06                           & 1.23                           & 0.12 \\
\rowcolor[rgb]{ .886,  .937,  .855} Ours (finetuned)               & \redxmark                      & \textbf{1.26}                  & \textbf{0.26}                  & \textbf{0.09} \\
\midrule
\rowcolor[rgb]{ .886,  .937,  .855} Ours                           & \greencheckmark                & 2.31                           & 1.52                           & 0.13 \\
\bottomrule
\end{tabular}%

}
\vspace{-8pt}
\caption{Results on ETH3D leaderboard (test set). All methods except for the last row have used ETH3D training set for fine-tuning. Our fine-tuned version ranks 1st on leaderboard at the time of submission. Last row is obtained via zero-shot inference from our foundation model.}
\label{tab:eth_finetune}
\vspace{-15pt}
\end{table}

%% file: table/ablation_foundation.tex
\begin{table}[h]
\centering
\def\mywidth{0.3\textwidth} 
\definecolor{green}{RGB}{0,200,0}
\resizebox{\mywidth}{!}{

\begin{tabular}{clc}
\toprule
Row                            & \multicolumn{1}{c}{Variations} & BP-2 \bigstrut\\
\midrule
1                              & DINOv2-L~\cite{oquab2023dinov2} & 2.46 \bigstrut[t]\\
2                              & DepthAnythingV2-S~\cite{yang2024depthanythingv2} & 2.22 \\
3                              & DepthAnythingV2-B~\cite{yang2024depthanythingv2} & 2.11 \\
4                              & \cellcolor[rgb]{ .886,  .937,  .855}DepthAnythingV2-L~\cite{yang2024depthanythingv2} & \cellcolor[rgb]{ .886,  .937,  .855}1.97 \bigstrut[b]\\
\cmidrule{2-3}5                              & STA (a)                        & 6.48 \bigstrut[t]\\
6                              & STA (b)                        & 2.22 \\
7                              & \cellcolor[rgb]{ .886,  .937,  .855}STA (c) & \cellcolor[rgb]{ .886,  .937,  .855}1.97 \bigstrut[b]\\
\cmidrule{2-3}8                              & Unfreeze ViT                   & 3.94 \bigstrut[t]\\
9                              & \cellcolor[rgb]{ .886,  .937,  .855}Freeze ViT & \cellcolor[rgb]{ .886,  .937,  .855}1.97 \bigstrut[b]\\
\bottomrule
\end{tabular}%
}
\vspace{-8pt}
\caption{Ablation study of STA module. Variations (a-c) correspond to Fig.~\ref{fig:sta}. The choices adopted in our full model are highlighted in green.}
\label{tab:sta_ablation}
\vspace{-10pt}
\end{table}

%% file: table/ablation_ahcf.tex
\begin{table}[h]
\centering
\def\mywidth{0.45\textwidth} 
\resizebox{\mywidth}{!}{
\begin{tabular}{clcrrrr}
\cmidrule{1-3}\cmidrule{5-7}Row                            & \multicolumn{1}{c}{Variations} & BP-2                           &                                & \multicolumn{1}{c}{Row}        & \multicolumn{1}{c}{Variations} & \multicolumn{1}{c}{BP-2} \\
\cmidrule{1-3}\cmidrule{5-7}1                              & RoPE                           & 2.19                           &                                & \multicolumn{1}{c}{10}         & \multicolumn{1}{l}{(3,3,1), (1,1,5)} & \multicolumn{1}{c}{2.10} \\
2                              & \cellcolor[rgb]{ .886,  .937,  .855}Cosine & \cellcolor[rgb]{ .886,  .937,  .855}1.97 &                                & \multicolumn{1}{c}{11}         & \multicolumn{1}{l}{(3,3,1), (1,1,9)} & \multicolumn{1}{c}{2.06} \\
\cmidrule{2-3}3                              & 1/32                           & 2.06                           &                                & \multicolumn{1}{c}{12}         & \multicolumn{1}{l}{(3,3,1), (1,1,13)} & \multicolumn{1}{c}{2.01} \\
4                              & \cellcolor[rgb]{ .886,  .937,  .855}1/16 & \cellcolor[rgb]{ .886,  .937,  .855}1.97 &                                & \multicolumn{1}{c}{13}         & \multicolumn{1}{l}{\cellcolor[rgb]{ .886,  .937,  .855}(3,3,1), (1,1,17)} & \multicolumn{1}{c}{\cellcolor[rgb]{ .886,  .937,  .855}1.97} \\
\cmidrule{2-3}5                              & Full                           & 2.25                           &                                & \multicolumn{1}{c}{14}         & \multicolumn{1}{l}{(3,3,1), (1,1,21)} & \multicolumn{1}{c}{1.98} \\
6                              & \cellcolor[rgb]{ .886,  .937,  .855}Disparity & \cellcolor[rgb]{ .886,  .937,  .855}1.97 &                                & \multicolumn{1}{c}{15}         & \multicolumn{1}{l}{(7,7,1), (1,1,17)} & \multicolumn{1}{c}{1.99} \\
\cmidrule{2-3}\cmidrule{5-7}7                              & Pre-hourglass                  & 2.06                           &                                &                                &                                &  \\
8                              & Post-hourglass                 & 2.20                           &                                &                                &                                &  \\
9                              & \cellcolor[rgb]{ .886,  .937,  .855}Parallel & \cellcolor[rgb]{ .886,  .937,  .855}1.97 &                                &                                &                                &  \\
\cmidrule{1-3}\end{tabular}%
}
\vspace{-8pt}
\caption{Ablation study of AHCF module. Left corresponds to DT, while right corresponds to APC. The choices adopted in our full model are highlighted in green.} \label{tab:ablation_ahcf}
\vspace{-15pt}
\end{table}

%% file: table/ablation_overall.tex
\begin{table}[t]
\centering
\def\mywidth{0.4\textwidth} 
\definecolor{green}{RGB}{0,200,0}
\resizebox{\mywidth}{!}{

\begin{tabular}{cccccrccr}
\cmidrule{1-5}\cmidrule{7-9}\multirow{2}[4]{*}{Row}        & \multirow{2}[4]{*}{STA}        & \multicolumn{2}{c}{AHCF}                                        & \multirow{2}[4]{*}{BP2}        &                                & Row                            & FSD                            & \multicolumn{1}{c}{BP2} \\
\cmidrule{3-4}\cmidrule{7-9}                               &                                & APC                            & DT                             &                                &                                & 1                              & \redxmark                      & \multicolumn{1}{c}{2.34} \\
\cmidrule{1-5}1                              &                                &                                &                                & 2.48                           &                                & 2                              & \cellcolor[rgb]{ .886,  .937,  .855}\greencheckmark & \multicolumn{1}{c}{\cellcolor[rgb]{ .886,  .937,  .855}1.15} \\
\cmidrule{7-9}2                              & \checkmark                     &                                &                                & 2.21                           &                                &                                &                                &  \\
3                              & \checkmark                     & \checkmark                     &                                & 2.16                           &                                &                                &                                &  \\
4                              & \checkmark                     &                                & \checkmark                     & 2.05                           &                                &                                &                                &  \\
5                              & \cellcolor[rgb]{ .886,  .937,  .855}\checkmark & \cellcolor[rgb]{ .886,  .937,  .855}\checkmark & \cellcolor[rgb]{ .886,  .937,  .855}\checkmark & \cellcolor[rgb]{ .886,  .937,  .855}1.97 &                                &                                &                                &  \\
\cmidrule{1-5}\end{tabular}%

}
\vspace{-0.1in}
\caption{\textbf{Left:} Ablation study of proposed network modules. \textbf{Right:} Ablation study of whether to use FSD dataset when training the foundation model described in Sec.~\ref{sec:implementation}. The choices adopted in our full model are highlighted in green.}
\label{tab:ablation_overall}
\vspace{-15pt}
\end{table}

%% file: suppl.tex
\clearpage
\setcounter{page}{1}
\maketitlesupplementary

\begin{figure*}[htb]
    \centering
    {\includegraphics[width=0.99\textwidth]{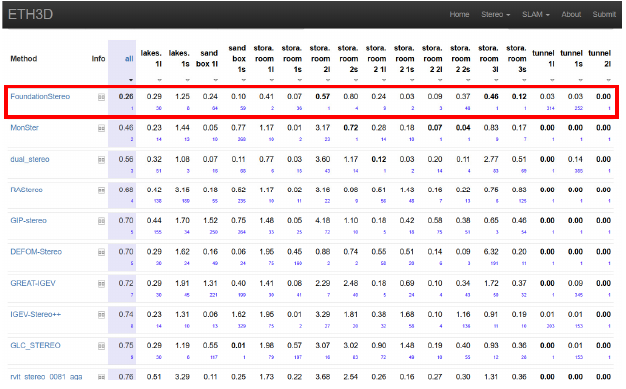}} 
    \vspace{-10pt}
    \caption{ETH3D leaderboard screenshot. Our fine-tuned foundation model (red box) ranks 1st at the time of submission.} 
    \label{fig:eth_leaderboard}
\end{figure*}

\section{ETH3D Leaderboard}
At the time of submission, our fine-tuned model ranks 1st on the \href{https://www.eth3d.net/low_res_two_view}{ETH3D leaderboard}, significantly outperforming both published and unpublished works. The screenshot is shown in Fig.~\ref{fig:eth_leaderboard}.

\begin{figure*}[h]
    \centering
    {\includegraphics[width=0.95\textwidth]{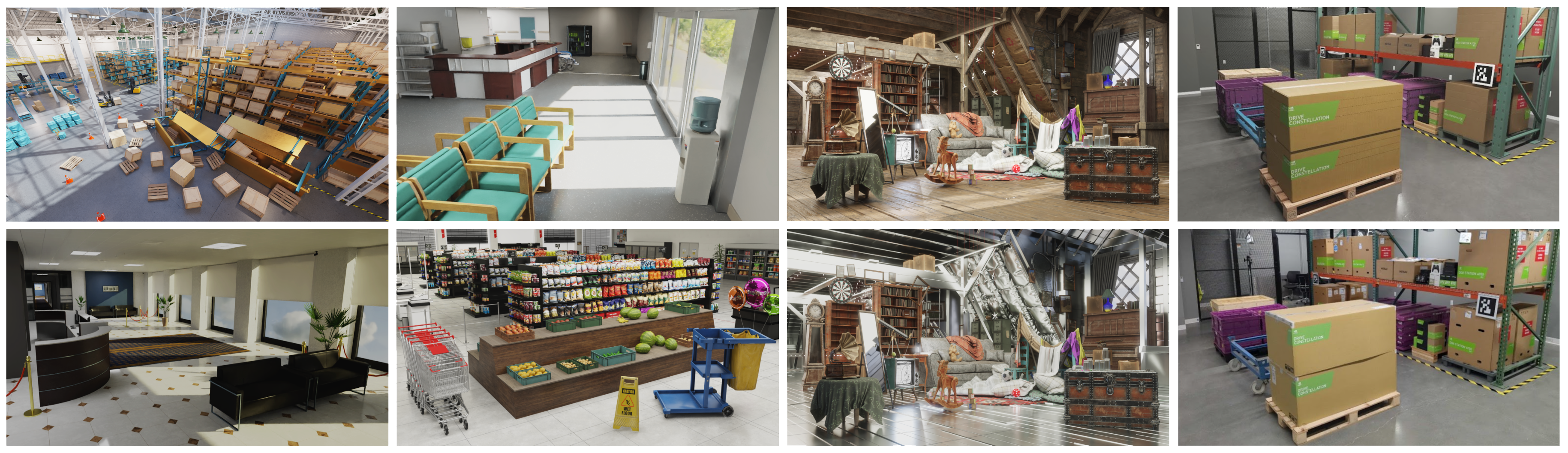}} 
    \vspace{-10pt}
    \caption{Examples scene models involving factory, hospital, wood attic, office, grocery store and warehouse. In the third column, we demonstrate an example of metallic material randomization being applied to augment scene diversity. The last column shows  comparison of a warehouse between the real (bottom) and our simulated digital twin (top) in high fidelity.} 
    \label{fig:scenario_model}
\end{figure*}

\section{Middlebury Leaderboard}
\begin{figure*}[htb]
    \centering
    {\includegraphics[width=0.99\textwidth]{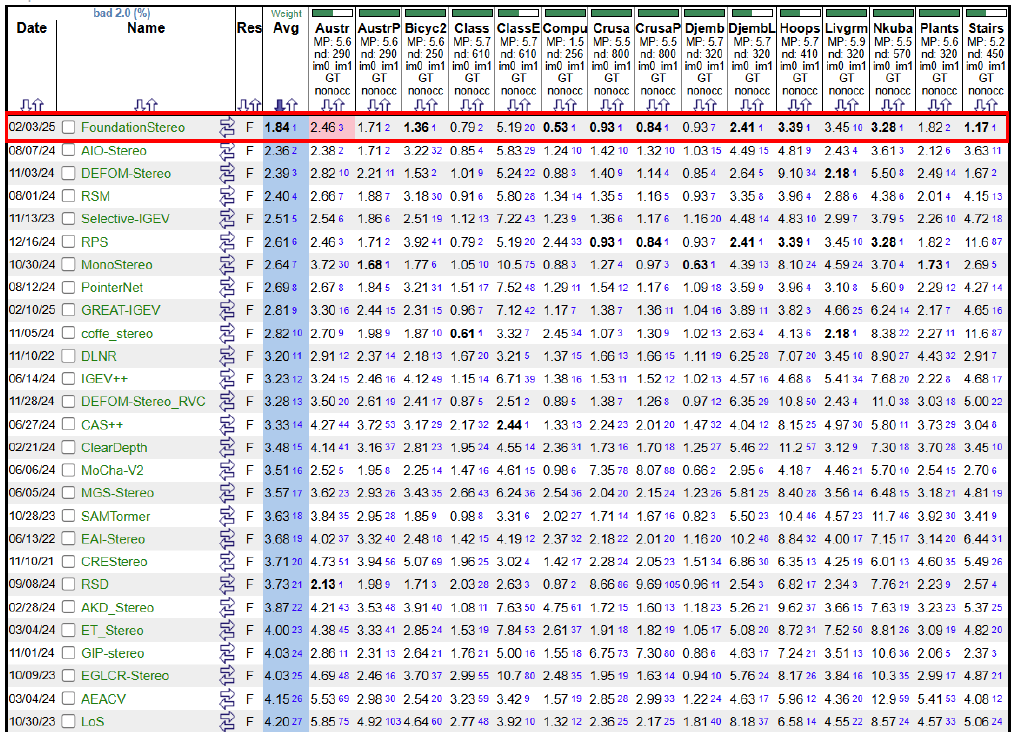}} 
    \vspace{-10pt}
    \caption{Middlebury leaderboard screenshot. Our fine-tuned foundation model (red box) ranks 1st at the time of submission.} 
    \label{fig:middlebury}
\end{figure*}

At the time of submission, our fine-tuned model ranks 1st on the \href{https://vision.middlebury.edu/stereo/eval3/}{Middlebury leaderboard}, significantly outperforming both published and unpublished works. The screenshot is shown in Fig.~\ref{fig:middlebury}.

\begin{figure*}[h]
    \centering
    {\includegraphics[width=0.6\textwidth]{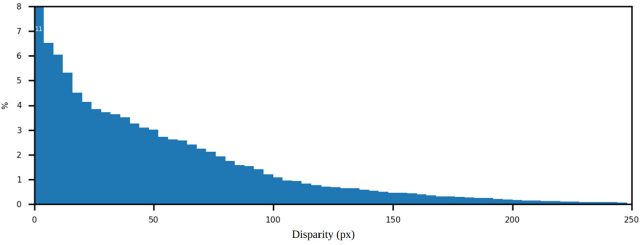}} 
    \vspace{-10pt}
    \caption{Disparity distribution in our proposed FSD.} 
    \label{fig:sdg_disp_range}
\end{figure*}

\section{More Ablation Study on Synthetic Data}
\boldparagraphstart{Effects of Self-Curation.} We study the effectiveness of self-curation pipeline introduced in Sec.~\ref{sec:sdg}. When disabling the self-curation while keeping the same data size, the synthetic dataset involves ambiguous samples that confuse the learning process, leading to slight performance drop when evaluated on Middlebury~\cite{middlebury} dataset.

\input{table/ablation_sdg_curation}

\boldparagraphstart{Effects of FSD for Other Methods.}  
Table~\ref{tab:zero_shot} (main paper) indicates benefits by introducing FSD for FoundationStereo model. To answer the question of whether FSD can benefit other methods beyond FoundationStereo, we now train representative works IGEV and Selective-IGEV on FSD and compare with their counterparts trained on Scene Flow. As shown in the table below, for both methods, our proposed FSD effectively boosts the performance compared to the commonly used Scene Flow dataset.

\input{table/sigev_vfm}

\section{Results on Translucent Objects}
We evaluate on Booster~\cite{ramirez2023booster} (half resolution), which is a challenging dataset consisting of specular and transparent objects. We compare with the most competitive methods from Fig.~\ref{fig:wild} (main paper) in the zero-shot setting. The quantitative and qualitative results are shown below.


\begin{minipage}{0.5\textwidth} 
  \begin{minipage}{0.4\textwidth} 
    \centering 
    \resizebox{\textwidth}{!}{\input{table/booster}} 
    \end{minipage} 
  \begin{minipage}{0.45\textwidth} 
    \centering     \includegraphics[width=\textwidth]{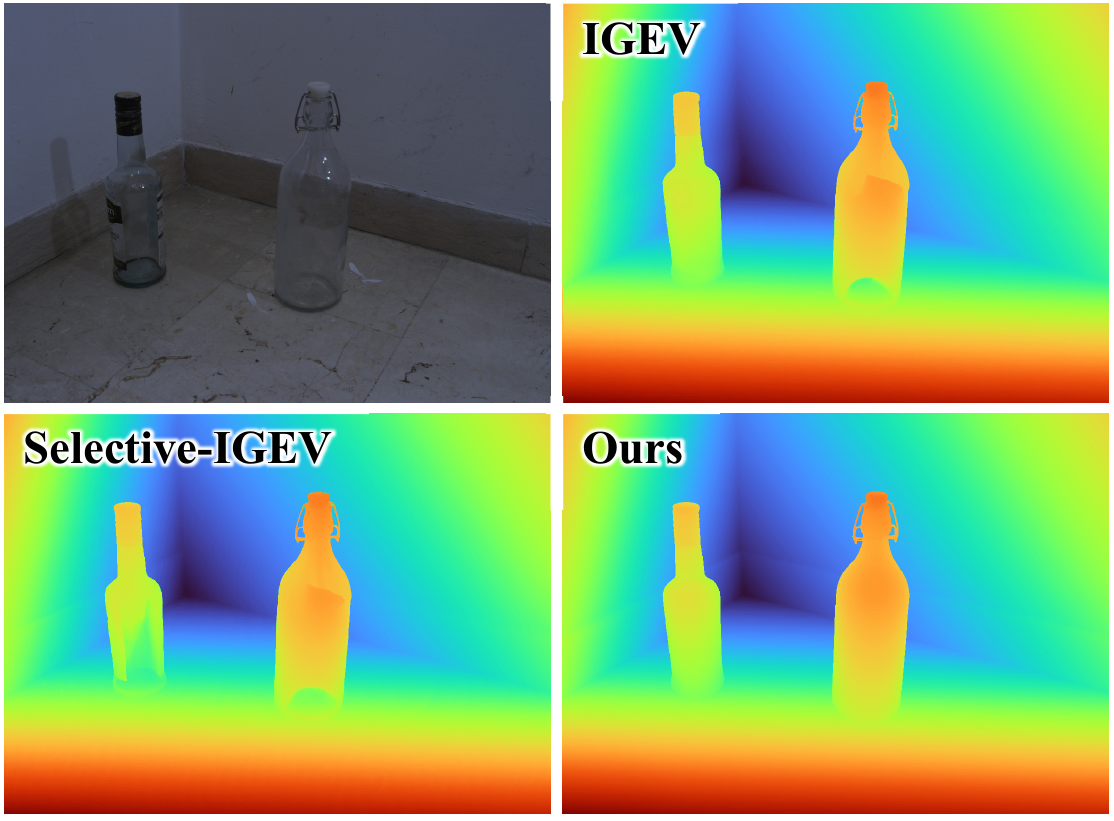}
  \end{minipage}   
\end{minipage}

\section{More Results on Middlebury Dataset}
We compare with competitive methods that released their public weights in zero-shot on Middlebury, shown in below table. Since NMRF~\cite{guan2024neural} did not report their evaluated Middlebury resolution, we rerun their released weights on all resolutions. At full resolution, maximum disparity 320 is used for FoundationStereo. Across all resolutions, ours significantly outperforms baselines. We also report the peak memory usage and running time averaged across the dataset on the same hardware, particularly single GPU 3090. On half and quarter resolutions, our peak memory occurs at STA module. On full resolution, it occurs at DT module. Despite the speed limitation which is not the focus when developing this work, ours can successfully run on a desktop GPU. Pruning or distillation remains an interesting future work to improve speed and memory footprint.

\input{table/middlebury_all_res}

\section{More Details of Synthetic Data Generation}

\boldparagraphstart{Tooling and Assets.}
The dataset generation is built on NVIDIA Omniverse. We use RTX path-tracing with 32 to 128 samples per pixel for high-fidelity photorealistic rendering. The data generation is performed across 48 NVIDIA A40 GPUs for 10 days. There are more than 5K object assets collected from varying sources including artist designs and 3D scanning with high-frequency geometry details. Object assets are divided into the groups of: furniture, open containers, vehicles, robots, floor tape, free-standing walls, stairs, plants, forklifts, dynamically animated digital humans, other obstacles and distractors. Each  group is defined with a separate randomization range for sampling locations, scales and appearances.  In addition, we curated 12 large scene models (Fig.~\ref{fig:scenario_model}), 16 skybox images, more than 150 materials, and 400 textures for tiled wrapping on object geometries for appearance augmentation. These textures are obtained from real-world photos and procedurally generated random patterns.

\boldparagraphstart{Camera Configuration.} For each data sample, we first randomly sample the stereo baseline camera focal length to diversify the  coverage of field-of-views and disparity distributions. Next, objects are spawned into the scene in two different methods to randomize the scene configuration: 1) camera is spawned in a random pose, and objects are added relative to the camera at random locations; 2) objects are spawned near a random location, and the camera is spawned nearby and oriented to the center of  mass of the object clutter. 

\boldparagraphstart{Layout Configuration.} We generate layouts in two kinds of styles: chaotic and realistic. Such combination of the more realistic structured layouts with the more randomized setups with flying objects has been shown to benefit sim-to-real generalization~\cite{tremblay2018deep}. Specifically, chaotic-style scenes involve large number of flying  distractors and simple scene layouts which consists of infinitely far skybox and a background plane. The lighting and object appearances (texture and material) are highly randomized. The realistic-style data uses indoor and outdoor scene models where the camera is restricted to locate at predefined areas. Object assets are dropped and applied with physical properties for collision. The simulation is performed randomly between 0.25 to 2 seconds to create physically realistic layouts with no penetration, involving both settled and falling objects. Materials and scales native to object assets are maintained and more natural lighting is applied. Among the realistic-style data, we further divide the scenes into three types which determine what categories of objects are selected to compose the scene for more consistent semantics:
\begin{myitem}
    \item Navigation - camera poses are often in parallel to the ground and objects are often spawned further away. Objects such as free-standing walls, furniture, and digital humans are sampled with higher probability.
    \item Driving - camera is often in parallel to the ground above the ground and objects are often spawned further away. Objects such as vehicles, digital humans, poles, signs and speed bumps are sampled with higher probability. 
    \item Manipulation - camera is oriented to face front or downward as in ego-centric views and objects are often spawned in closer range to resemble interaction scenarios. Objects such as household or grocery items, open containers, robotic arms are sampled with higher probability. 
\end{myitem}

\boldparagraphstart{Lighting Configuration.} Light types include global illumination, directed sky rays, lights baked-into 3D scanned assets, and light spheres which add dynamic lighting when spawned near to surfaces. Light colors, intensities and directions are randomized. Lighting vibes such as daytime, dusk and night are included within the random sampling ranges.

\boldparagraphstart{Disparity Distribution.} Fig.~\ref{fig:sdg_disp_range} shows the disparity distribution of our FSD dataset.

\section{Acknowledgement} 
We would like to thank Gordon Grigor, Jack Zhang, Xutong Ren, Karsten Patzwaldt, Hammad Mazhar and other NVIDIA Isaac team members for their tremendous engineering support and valuable discussions.

%% file: table/ablation_sdg_curation.tex
\begin{table}[h]
\centering
\def\mywidth{0.2\textwidth} 
\resizebox{\mywidth}{!}{

\begin{tabular}{cc}
\toprule
Variation                      & BP2 \\
\midrule
\rowcolor[rgb]{ .886,  .937,  .855} W/ self-curation               & 1.15 \\
W/o self-curation              & 1.27 \\
\bottomrule
\end{tabular}%

}
\vspace{-8pt}
\caption{Effectiveness of self-curation pipeline when generating synthetic data.} \label{tab:ablation_sdg_curation}
\end{table}

%% file: table/sigev_vfm.tex
\begin{table}[h]
\centering
\def\mywidth{0.49\textwidth} 
\definecolor{green}{RGB}{0,200,0}
\resizebox{\mywidth}{!}{

\begin{tabular}{l|c|rrrr}
\thickline
\multirow{2}[2]{*}{Methods}    & \multirow{2}[2]{*}{Train data} & Middlebury                     &  ETH3D                         & KITTI-12                       & KITTI-15  \bigstrut[t]\\
                               &                                & BP-2                           & BP-1                           & D1                             & D1 \bigstrut[b]\\
\hline
IGEV                           & Scene Flow                     & 8.8                            & 4.0                            & 5.2                            & 5.7 \bigstrut[t]\\
IGEV                           & FSD                            & 7.8                            & 3.5                            & 3.2                            & 4.7 \\
Selective-IGEV                 & Scene Flow                     & 9.2                            & 5.7                            & 4.5                            & 5.6 \\
Selective-IGEV                 & FSD                            & 7.9                            & 3.5                            & 3.0                            & 4.4 \bigstrut[b]\\
\thickline
\end{tabular}%

}
\vspace{-10pt}
\caption{Effects of FSD for other methods.}\label{tab:fsd_effect}

\end{table}

%% file: table/booster.tex

\begin{tabular}{l|rrrr}
\hline
\multirow{2}[2]{*}{Methods}    & \multicolumn{4}{c}{Half} \bigstrut[t]\\
                               & BP1                            & BP-2                           & BP-3                           & EPE \bigstrut[b]\\
\hline
Selective-IGEV                 & 23.8                           & 15.0                           & 12.0                           & 6.6 \bigstrut[t]\\
IGEV                           & 30.8                           & 22.3                           & 19.0                           & 22.7 \\
\rowcolor[rgb]{ .886,  .937,  .851} Ours                           & \textbf{19.0}                  & \textbf{9.6}                   & \textbf{6.7}                   & \textbf{2.2} \bigstrut[b]\\
\hline
\end{tabular}%

%% file: table/middlebury_all_res.tex
\begin{table}[h]
\centering
\def\mywidth{0.49\textwidth} 
\definecolor{green}{RGB}{0,200,0}
\resizebox{\mywidth}{!}{

\begin{tabular}{c|rrr|rrr|rrr}
\thickline
\multirow{2}[2]{*}{Methods}    & \multicolumn{3}{c|}{Full }                                                                       & \multicolumn{3}{c|}{Half}                                                                        & \multicolumn{3}{c}{Quarter} \bigstrut[t]\\
                               & BP-2                           & \makecell{peak\\ mem (G)}      & time (s)                       & BP-2                           & \makecell{peak\\ mem (G)}      & time (s)                       & BP-2                           & \makecell{peak\\ mem (G)}      & time (s) \bigstrut[b]\\
\hline
\makecell{Selective-\\IGEV[61]} & 12.9                           & 6.9                            & 2.52                           & 9.2                            & 1.7                            & 0.72                           & 7.0                            & 0.5                            & 0.25 \bigstrut[t]\\
IGEV[33]                       & 13.1                           & 6.3                            & 2.06                           & 8.8                            & 1.6                            & 0.53                           & 6.4                            & 0.5                            & 0.18 \\
IGEV++[66]                     & 12.7                           & 13.1                           & 2.12                           & 7.8                            & 3.4                            & 0.50                           & 6.3                            & 0.9                            & 0.15 \\
NMRF[20]                       & 35.3                           & 8.1                            & 0.95                           & 10.9                           & 1.8                            & 0.20                           & 5.0                            & 0.5                            & 0.05 \\
\rowcolor[rgb]{ .886,  .937,  .851} Ours                           & \cellcolor[rgb]{ .886,  .937,  .855}\textbf{4.8} & \cellcolor[rgb]{ .886,  .937,  .855}18.5 & \cellcolor[rgb]{ .886,  .937,  .855}8.14 & \cellcolor[rgb]{ .886,  .937,  .855}\textbf{1.1} & \cellcolor[rgb]{ .886,  .937,  .855}10.5 & \cellcolor[rgb]{ .886,  .937,  .855}2.97 & \cellcolor[rgb]{ .886,  .937,  .855}\textbf{1.3} & \cellcolor[rgb]{ .886,  .937,  .855}2.3 & \cellcolor[rgb]{ .886,  .937,  .855}0.55 \bigstrut[b]\\
\thickline
\end{tabular}%

}
\vspace{-10pt}
\caption{Results on varying resolutions in Middlebury.}\label{tab:middlebury}
\end{table}